\documentclass[11pt]{article}
\usepackage{coling2020}
\usepackage{times}
\usepackage{url}
\usepackage{latexsym}

\usepackage{graphicx}
\usepackage{multirow}

\colingfinalcopy % Uncomment this line for the final submission

\title{Residual Tree Aggregation of Layers for Neural Machine Translation}

\author{GuoLiang Li $^{1}$ , Yiyang Li $^{2}$  \\
	{\tt }\\
$^{1}$ \quad KingSoft AI Lab \\
$^{2}$ \quad Massey University}

\pdfoutput=1

\begin{document}
\maketitle
\begin{abstract}
Although attention-based Neural Machine Translation has achieved remarkable progress in recent layers, it still suffers from issue of making insufficient use of the output of each layer. In transformer, it only uses the top layer of encoder and decoder in the subsequent process, which makes it impossible to take advantage of the useful information in other layers. To address this issue, we propose a residual tree aggregation of layers for Transformer(RTAL), which helps to fuse information across layers. Specifically, we try to fuse the information across layers by constructing a post-order binary tree. In additional to the last node, we add the residual connection to the process of generating child nodes. Our model is based on the Neural Machine Translation model Transformer and we conduct our experiments on WMT14 English-to-German and WMT17 English-to-France translation tasks. Experimental results across language pairs show that the proposed approach outperforms the strong baseline model significantly.
\end{abstract}

\section{Introduction}
\label{intro}

%
% The following footnote without marker is needed for the camera-ready
% version of the paper.
% Comment out the instructions (first text) and uncomment the 8 lines
% under "final paper" for your variant of English.
%
%\blfootnote{
    %
    % for review submission
    %
  %  \hspace{-0.65cm}  % space normally used by the marker
    %*Work was done  in 2018 when GuoLiang Li and Yiyang Li was working at KingSoft AI Lab.
    %
    % % final paper: en-uk version
    %
    % \hspace{-0.65cm}  % space normally used by the marker
    % This work is licensed under a Creative Commons
    % Attribution 4.0 International Licence.
    % Licence details:
    % \url{http://creativecommons.org/licenses/by/4.0/}.
    %
    % % final paper: en-us version
    %
    % \hspace{-0.65cm}  % space normally used by the marker
    % This work is licensed under a Creative Commons
    % Attribution 4.0 International License.
    % License details:
    % \url{http://creativecommons.org/licenses/by/4.0/}.
%}
Due to the excellent performance, deep neural networks have been widely used in computer vision and nature language processing tasks. A stack of many layers is the obvious feature for deep neural networks, and how to transform and fuse information across layers efficiently is a key challenge. Neural Machine Translation(NMT), which is a multi-layer end-to-end structure, has achieved state-of-the-art performances in large-scale translation tasks \cite{bahdanau2014neural,luong2015effective}. More recently, the system based on self-attention \cite{vaswani2017attention} is rapidly becoming the standard component in NMT, it demonstrates both superior performance and training speed compared to previous architectures using recurrent neural network \cite{wu2016google}.

For vanilla Transformer, it only leverages the top layers of encoder and decoder in the subsequent process, which makes it impossible to take advantage of the useful information in other layers. At the same time, in the work of \cite{shi2016does} show that both local and global syntactic information about source sentences is captured by the encoder, and different types of syntax is stored in different layers. Fusing the outputs of different layers is helpful to improve the model performance, some promising attempts have proven to be of profound value in computer vision tasks \cite{yu2018deep} and nature language process tasks \cite{shen2018dense,dou2018exploiting}.
In this work, we continue the line of research and go towards a more efficient approach to fuse information across layers. We try to fuse information across layers through residual tree aggregation of layers, it consists of a post-order binary tree and residual connection. To achieve better model performance, we validate many kinds of aggregation formulas as well. Our contributions are threefold:

\begin{itemize}
	\item Inspired by the approaches of dealing with the output of each layer in numerical analysis \cite{peters2018deep,he2016deep,huang2017densely}, we propose an approach based on residual tree aggregation of layers(RTAL) to fuse information across layers. This overcomes the problem with the standard NMT network where only uses the top layer of encoder and decoder, and the residual connection in RTAL contributes to enrich each child node of the binary tree.
	\item The proposed approach is efficient and easy to implement, and it also can be used in arbitrary multi-layer deep neural networks.
	\item Our work is among the few studies \cite{dou2019dynamic,shen2018dense,wang2018multi} which prove that the idea of fusing information across layers can have promising applications on natural language processing tasks.
\end{itemize}
We evaluate our approach on WMT14 English-to-German(En-De) and WMT17 English-to-France(En-Fr) translation tasks, and we employ the Transformer \cite{vaswani2017attention}as the strong baseline model. Experimental results show that our approach contributes to make full use of the output of each layer in Transformer across language pairs, it outperforms the Transformer-Big model by 1.19 BLEU points on En-De translation task and 0.35 BLEU points on En-Fr translation task. In addition, our proposed approach is robust, it always outperforms the baseline model when we use different aggregation formulas.

\section{Related Work}
Training neural networks with multiple stacked layers is challenging, making full use of the outputs of different layers can drastically improve the model performance from computer vision tasks to nature language processing tasks. \cite{he2016deep} propose a residual learning framework, which combine layers and encourage gradient flow by simple shortcut connections. \cite{yu2018deep} augment standard architectures with deeper aggregation to better fuse information, and designs architectures iteratively and hierarchically merge the feature hierarchy to make networks with better accuracy and fewer parameters.

Concerning nature language processing tasks, \cite{shen2018dense} propose a densely connected NMT architecture, it not only is able to use features from all previous layers to create new features for both encoder and decoder, but also uses the dense attention structure to improve attention quality. \cite{dou2019dynamic} propose to use routing-by-agreement strategies to aggregate layers dynamically, the algorithm learns the probability of individual layer representations assigned to aggregated representations in an iterative way. \cite{he2018layer} propose the concept of layer-wise coordination for NMT, which explicitly coordinates the learning of hidden representations of the encoder and decoder together layer by layer, gradually from low level to high level.

Although existing works have achieved remarkable progress in making full use of the output of each layer, there are still some shortcomings. In the work of \cite{shen2018dense,dou2019dynamic}, they modify the encoder and decoder of the original Transformer by adding extra layers, which increases the size of model and slows down the speed of inference. In the work of \cite{he2018layer}, it keeps the similar size of the model, but it breaks the origin structure of each layer in Transformer. In the case of ensuring that the original structure of each layer is unchanged, this approach can not be applied to arbitrary multi-layer deep neural networks. In our work, we propose a residual tree aggregation of layers for NMT, which is not suffered from the above issues.

\section{Background}
\subsection{Neural Machine Translation}
The encoder-decoder framework has noticeable effect on neural machine translation tasks \cite{sutskever2014sequence,wu2016google}. Generally, multi-layer encoder and decoder are employed to perform the translation task through a series of nonlinear transformations. The encoder reads the source sentence denoted by $X=(x_{1}, x_{2}, ..., x_{M})$ and maps it to a continuous representation $Z=(z_{1}, z_{2}, ..., z_{M})$. Given $Z$, the decoder generates the target translation $Y=(y_{1}, y_{2}, ..., y_{N})$ conditioned on the sequence of tokens previously generated. The encoder-decoder framework model learns its parameter $\theta$ by maximizing the log-likelihood $p(Y|X;\theta)$, which is usually decomposed into the product of the conditional probability of each target word.
\begin{equation}
p(Y|X;\theta) = \prod_{t=1}^{N}{p(y_{t}|y_{1},...,y_{t-1},X;\theta)}
\end{equation}
where $N$ is the length of target sentence. Both the encoder and decoder can be implemented by different structure of neural models, such as RNN \cite{cho2014learning}, CNN \cite{gehring2017convolutional} and self-attention \cite{vaswani2017attention}.
\subsection{Self-attention based network}
Self-attention networks \cite{shaw2018self,so2019evolved} have attracted increasing attention due to their flexibility in parallel computation and dependency modeling. Self-attention networks calculate attention weights between each pair of tokens in a single sequence, and then a position-wise feed-forward network to increase the non-linearity. The self-attention is formulated as:
\begin{equation}
Attention(Q,K,V) = softmax(\frac{QK^{T}}{\sqrt{d_{model}}})V
\end{equation}
where $Q,K,V$ denotes the query, key and value vectors,respectively, $d_{model}$ is the dimension of hidden representations. The position-wise feed-forward network consists of a two-layer linear transformation with ReLU activation in between:
\begin{equation}
FFN(x) = max(0, xW_{1} + b_{1})W_{2} + b_{2}
\end{equation}
Instead of performing a single attention function, it is beneficial to capture different context features with multiple individual attention functions, namely multi-head attention. Specifically, multi-head attention model first transforms $Q,K$ and $V$ into $H$ subspaces with different linear projections:
\begin{equation}
Q_{h}, K_{h}, V_{h} = QW_{h}^{Q}, KW_{h}^{K}, VW_{h}^{V}
\end{equation}
where $Q_{h}, K_{h}, V_{h}$ are the query, key and value representation of the $h$ head,respectively. $W_{h}^{Q},W_{h}^{K},W_{h}^{V}$ denote parameter matrices associated with the $h$ head. $H$ attention functions are applied in parallel to produce the output states $(o_{1}, o_{2}, ..., o_{H})$, then the $H$ output states are concatenated and linearly transformed to produce the final state. The process is formulated as:
\begin{equation}
o_{h} = Attention(Q_{h}, K_{h}, V_{h})
\end{equation}
\begin{equation}
O = [o_{1}, o_{2}, ..., o_{H}]W^{o}
\end{equation}
where $O$ is the final output states, and $W^{o}$ is a trainable matrix.

\section{Approach}
In this section, we first introduce the idea of RTAL, and then we also describe the aggregation formulas used in our experiments.
\subsection{Residual Tree Aggregation of Layers}
\begin{figure*}[t]
	\centering
	\includegraphics[width=0.7\textwidth]{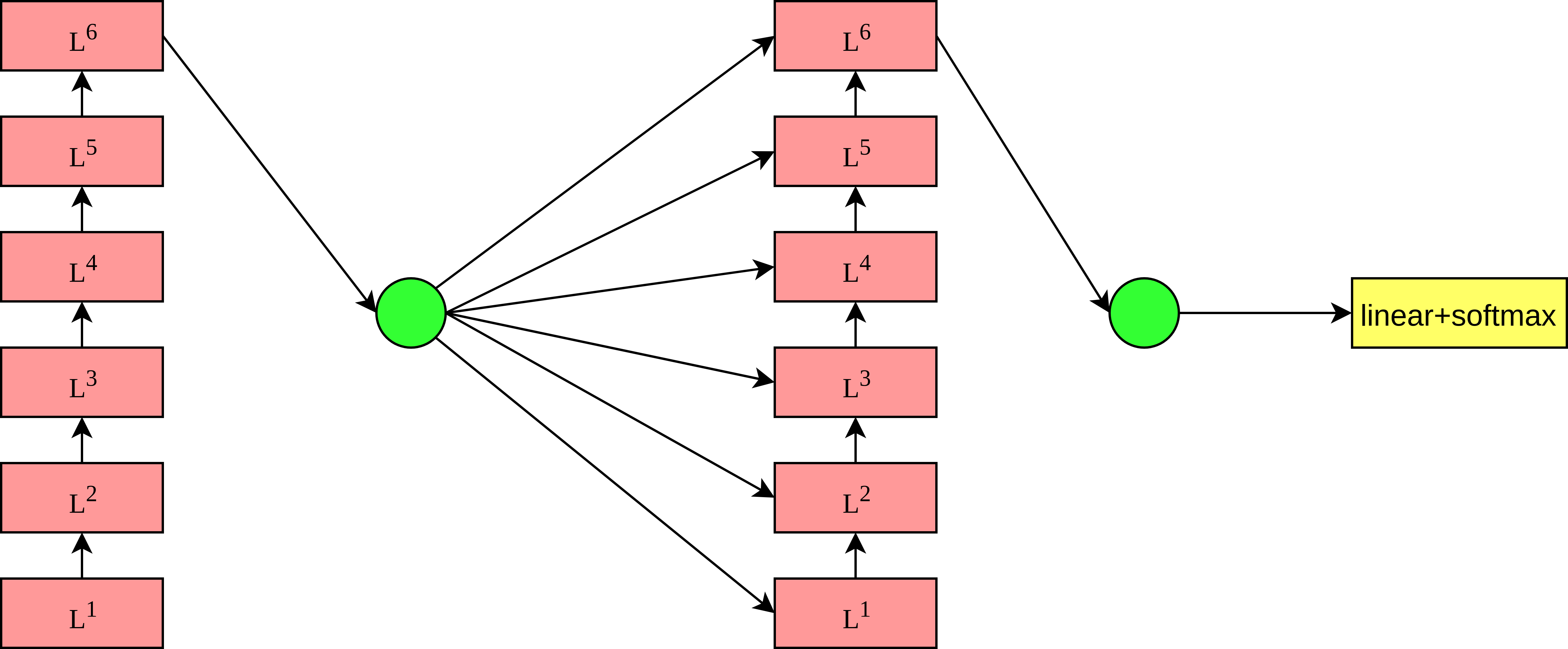} % Reduce the figure size so that it is slightly narrower than the column.
	\caption{The Transformer-model architecture. The rectangle denotes a layer in the encoder or decoder, the circle denotes the outputs of the last layer in the encoder and decoder. The output of the last layer in the encoder is used as the input of each layer in the decoder, the output of the last layer in the decoder is used to generate the output sequence.}
	\label{fig1}
\end{figure*}
\begin{figure*}[t]
	\centering
	\includegraphics[width=0.8\textwidth]{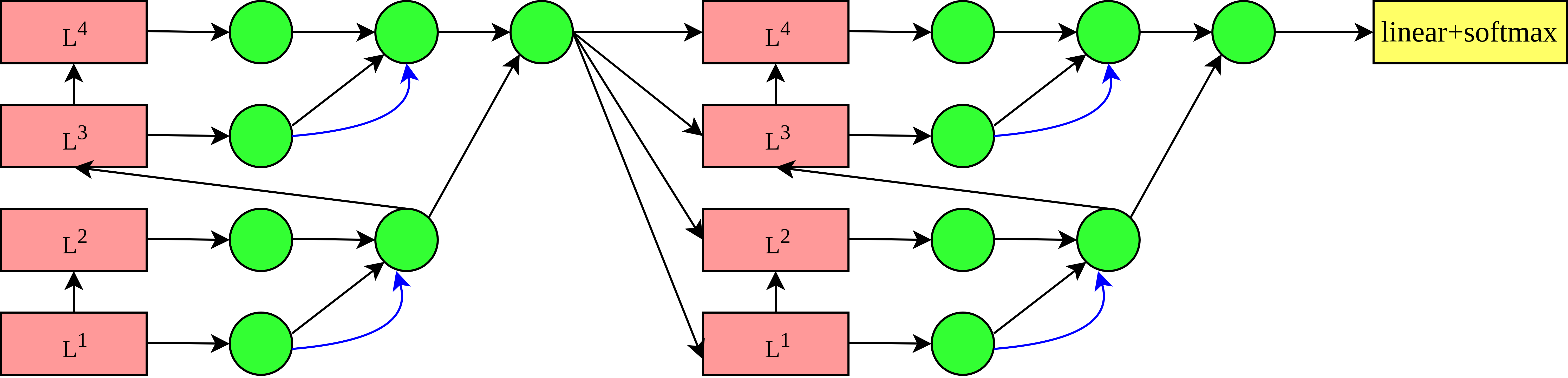} % Reduce the figure size so that it is slightly narrower than the column.
	\caption{The RTAL based on Transformer, where the encoder and decoder are both composed of stack of 4 layers. The RTAL consists of a post-order binary tree and residual connection. The blue line denotes the residual connection.}
	\label{fig2}
\end{figure*}
As the figure \ref{fig1} shows, the Transformer only uses the top layer of the encoder and decoder in the subsequent process, which makes it impossible to take advantage of the useful information in other layers. To address this issue, we propose a residual tree aggregation of layers for Transformer(RTAL), it contributes to fuse information across layers. In principle, the RTAL can be applied to any multi-layer neural models, such as Convs2s \cite{gehring2017convolutional} and Transformer \cite{vaswani2017attention}. In this work, we directly focus on Transformer considering that it achieves excellent performance on many translation tasks.

RTAL keeps the original structure of Transformer unchanged, it only add extra layers in the encoder and decoder. As the figure \ref{fig3} shows, RTAL are used to make full use of the output of each layer in encoder and decoder, it consists of a post-order binary tree and residual connection. Specifically, we try to fuse information by constructing a post-order binary tree. In the process of generating child nodes, the residual connection is also used except for the last node. The post-order binary tree implements hierarchical aggregation, and the residual connection provides extra information for generating child nodes. In other computer vision and nature language processing tasks, hierarchical aggregation and residual connection have also proven to be of profound value \cite{yu2018deep,dou2018exploiting,he2016deep}.

In Transformer, the encoder and decoder are both composed of stack of 6 layers. RTAL requires the number of layer is $2^{n}$, although RTAL can be applied to the last 4 layers in Transformer and improves the model performance significantly, it also increases the size of model, these will be described in our experiments. Under the constrain of model size, we try to reduce the number of layer in the encoder and decoder and set them to be 4. This approach reduces the size of model effectively, and the loss of model performance is small.

\subsection{Aggregation formulas}
\begin{figure}[t]
	\centering
	\includegraphics[width=0.15\textwidth]{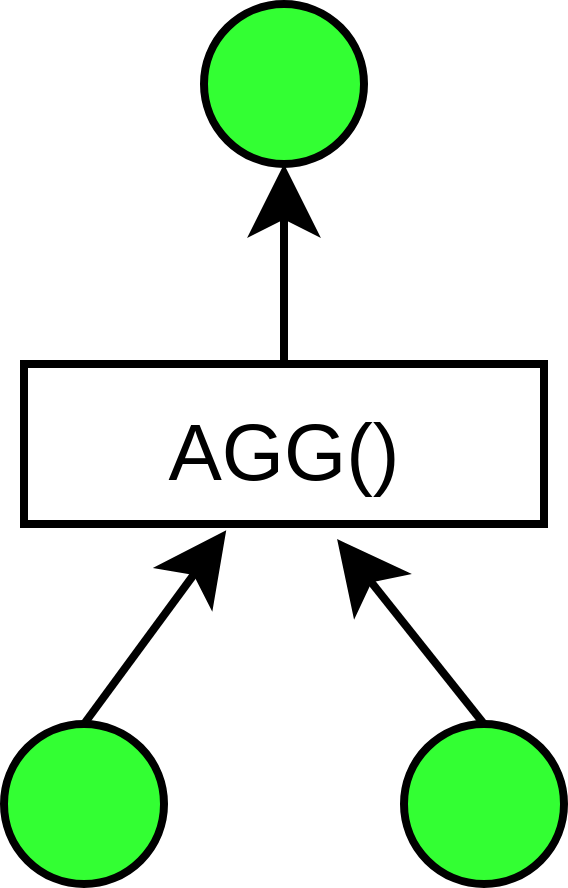} % Reduce the figure size so that it is slightly narrower than the column.
	\caption{Generating the child nodes by aggregating the layer outputs, the circles denote layer outputs and child nodes, $AGG()$ denotes the aggregation formula.}
	\label{fig3}
\end{figure}

In order to get better model performance, we try many kinds of aggregation formulas in our RTAL. The aggregation formula is used to generate child nodes, the details are shown in figure \ref{fig3}.

\subsubsection{Aggregation by averaging}
A simple way of aggregating all hidden representations is to average them. Given the layer outputs$(h_{i}, h_{i+1})$, the output of average-based aggregation is formulated as:
\begin{equation}
AGG(h_{i}, h_{i+1}) = 0.5 * \sum(h_{i},h_{i+1})
\end{equation}
This method does not add extra parameters and simply assigns an equal weight to the layer outputs. It is easy to implement and has little effect on model training speed.
\subsubsection{Aggregation by concatenation}
A more common method is to use concatenation operation and feed-forward neural network(FFN) for fusion. Specifically, we first concatenate the layer outputs$(h_{i}, h_{i+1})$ to form a bigger vector. Then, we feed the bigger vector into the FFN layer, which only has a single hidden layer. Let $d$ be the size of the layer output vector, the formulated representation of this method is defined as:
\begin{equation}
AGG(h_{i}, h_{i+1}) = FFN(concat(h_{i},h_{i+1}))
\end{equation}

Where $h_{i}$ and $AGG(h_{i}, h_{i+1})$ are both $d$ dimensional vectors, the concatenation of $h_{i}$ and $h_{i+1}$ is a $2*d$ dimensional vector. The FFN transforms the $2*d$ dimensional vector to a $d$ dimensional vector, and the non-linear activation functions used in FFN contributes to obtain a better result \cite{wang2018multi}. The concatenation between layer outputs generates bigger vectors, and it takes more expensive computation in the subsequent process.
\subsubsection{Aggregation by element-wise production}
In addition to the averaging and concatenation-based methods, we design a simple and efficient aggregation functions as well. Since the concatenation operation will generate a bigger dimensional vector, we propose to use element-wise production to deal with the layer outputs and keep the dimension of the vector unchanged. At the same time, the FFN layer is also used to transform the vector generated by point-wise production. We formulate the process like this:
\begin{equation}
sumb = (h_{i} + h_{i+1}) * \beta\label{ewp:a}
\end{equation}
\begin{equation}
AGG(h_{i}, h_{i+1}) = FFN(LN(sumb)) + sumb\label{ewp:b}
\end{equation}

Where $LN()$ denotes the layer normalization \cite{lei2016layer}, $\beta$ is a hyper-parameter and is trainable. Dropout \cite{srivastava2014dropout} and residual connection \cite{he2016deep} are also employed.

\section{Experiment}

\subsection{Setup}
\subsubsection{Datasets and Evaluation}
To make the results more convincing, we evaluate our approach on WMT14 English-to-German(En-De) and WMT17 English-to-France(En-Fr) translation tasks.

On the En-De translation task, we use the same data set with \cite{vaswani2017attention}'s work. The dataset contains about 4.5 million sentence pairs, with 116 million English words and 110 million German works. On the En-Fr translation task, we use the significantly larger dataset consisting of 40 million sentence pairs. All the data has been tokenized and jointly byte pair encoded\cite{sennrich2015neural} with 32k merge operations using a shared vocabulary. We use newstest2013 for validation set and newstest2014 for test set.

For evaluation, we average the last 5 checkpoints for the base model and average the last 20 checkpoints for the big model. We use beam search with a beam size of 4 and length penalty $\alpha=0.6$. We measure case-sensitive tokenized BLEU for En-De and En-Fr translation tasks.
\subsubsection{Model and Hyperparameters}
All models were trained on the tensor2tensor \cite{tensor2tensor} with 8 NVIDIA P100 GPUs, where each was allocated with a batch size of 4096 tokens. For our base model, the hidden dimension $d_{x}=512$, the number of attention heads are 8, and the dimension of feed-forward inner-layer is 2048. For our big model, the size of hidden dimension, the number of attention heads and the dimension of feed-forward inner-layer are double. The base model was trained for 100k steps and the big model was trained for 300k steps. All models were optimized by Adam \cite{kingma2014adam} with $\beta_{1}=0.9, \beta_{2}=0.98$ and $\epsilon=10^{-9}$, the label smoothing $\epsilon_{ls}=0.1$ was used as regularization. In generally, we followed the configure as suggested in tensor2tensor, we employ the Transformer as the baseline model.
\subsection{Results}
\subsubsection{Results on the En-De task}
In table \ref{table1}, we first show the results of the En-De translation task, which we compare to the existing systems based on self-attention. Obviously, most of the model variants outperform the Transformer significantly, although they have a larger amount of parameters. At the same time, the big-model usually has better model performance than the base-model, and it also requires more computational cost.

As for our approach, we first verify the idea of RTAL on the Transformer-Base and Transformer-Big models, where both the encoder and decoder are composed of stack of 6 layers. The results show that they outperform the Transformer baseline significantly and have 1.16 and 1.19 BLEU points higher than baseline models. Compared to the work of \cite{dou2018exploiting}, the RTAL based on the Transformer-Big model has fewer parameters and higher BLEU points, the RTAL based on the Transformer-Base model has similar BLEU ascent and fewer parameters. Compared to the work of \cite{he2018layer}, the RTAL based on the Transformer-Base and Transformer-Big models have higher BLEU points.
\begin{table*}[t]
\begin{center}
\begin{tabular}{ccccc}
System & Architecture & Param & BLEU & $\Delta$ \\
\hline
\cite{wu2016google}& GNMT & - & 26.30 & -\\
\cite{gehring2017convolutional} & ConvS2S & - & 26.36 & - \\
\multirow{2}*{\cite{vaswani2017attention}} & Transformer-Base & 65M & 27.3 & - \\
& Transformer-Big & 213M & 28.4 & - \\
\multirow{2}*{\cite{shaw2018self}} & Transformer-Base + Relative Position & - & 26.8 & -0.5 \\
& Transformer-Big + Relative Position & $\dagger$210M & 29.2 & +0.8 \\
\multirow{2}*{\cite{dou2018exploiting}} & Transformer-Base + Deep Representations & 111M & 28.78 & +1.14$\ddagger$ \\
& Transformer-Big + Deep Representations & 356M & 29.2 & +0.63$\ddagger$ \\
\multirow{2}*{\cite{he2018layer}} & Layer-wise Coordination-Base & - & 28.3 & +1.0 \\
& Layer-wise Coordination-Big & $\dagger$210M & 29.0 & +0.6 \\
\hline
\multirow{4}*{Our work} & Transformer-Base + RTAL (4L) & 59M & 27.3 & +0.0 \\
& Transformer-Big + RTAL (4L)& 202M & 29.28 & +0.88 \\
& Transformer-Base + RTAL (6L)& 73M & 28.46 & +1.16 \\
& Transformer-Big + RTAL (6L)& 261M & 29.58 & +1.19 	
\end{tabular}
\end{center}
\caption{\label{table1} Comparing with existing NMT systems on WMT14 English-to-German translation task. "(4L)" denotes the encoder and decoder are both composed of stack of 4 layers. $\dagger$ denotes an estimate value. $\ddagger$ denotes that this model has another baseline model. "Param" denotes the trainable parameter size of each model(M=million).}\smallskip
\end{table*}

Although our approach contributes to get better model performance, it increases the size of model compared to the Transformer baseline model. To reduce the size of model, we also train the RTAL based on the Transformer variants, where both the encoder and decoder are composed of stack of 4 layers. As we can see, the RTAL based on the Transformer-Big variant have similar size of model compared to the Transformer-Big model, it still outperforms the Transformer-Big baseline model by 0.98 BLEU points. At the same time, it also outperforms the Big-model in the work of \cite{dou2018exploiting,he2018layer} and has fewer parameters.
\subsubsection{Results on the En-Fr task}
Seen from the En-De task, the RTAL based on the Transformer is more effective, where both the encoder and decoder have stack of 6 layers. Therefore, we evaluate our approach with the same structure on the En-Fr task. As seen in table \ref{table2}, we retrain the Transformer-Base and Transformer-Big models, and our approach outperforms the baseline models by 0.38 and 0.35 BLEU points. It confirms that our approach is a promising strategy for fusing information across layers in Transformer.
\begin{table}[t]
\begin{center}
\begin{tabular}{cccc}
System & Architecture & BLEU & $\Delta$ \\
\hline
\multirow{4}*{Our work} & T-Base & 41.26 & - \\
& T-Big & 42.58 & - \\
& T-Base + RTAL(6L) & 41.64 & +0.38 \\
& T-Big + RTAL(6L) & 42.93 & +0.35
\end{tabular}
\caption{\label{table2} Experimental results on WMT17 English-to-France translation task. "T-Base" denotes the Transformer-Base model. "T-Big" denotes the Transformer-Big model. "(6L)" denotes the encoder and decoder are both composed of stack of 6 layers.}\smallskip
\end{center}
\end{table}
\subsection{Analysis}
We conducted extensive analysis from different perspectives to better understand our model. All results are reported on the En-De translation task with Base-Model parameters, where the encoder and decoder are both composed of stack of 6 layers.
\subsubsection{Effect on Encoder and Decoder}
Both the encoder and decoder are multiple stacked layers, which may benefit from our proposed approach. In this experiment, we investigate the effects of RTAL on the encoder and decoder.
\begin{table}[h]
	\begin{center}
		\begin{tabular}{ccc}
			System & Aggregation position & BLEU \\
			\hline
			\multirow{3}*{T-Base + RTAL(6L)} & +encoder & 27.77 \\
			& +decoder & 27.82 \\
			& +both & 28.46
		\end{tabular}
		\caption{\label{table3} Experimental results of applying RTAL to different components on En-De test set. "+encoder" denotes that  RTAL is applied to the encoder. "+both" denotes that RTAL is applied to the encoder and decoder.}\smallskip
	\end{center}
\end{table}

As shown in table \ref{table3}, RTAL used in the encoder or decoder individually consistently outperforms the Transformer baseline model, and it further improves the model performance when using RTAL simultaneously in the encoder and decoder. These results claim that fusing information across layers is useful for both encoding the input sequence and generating output sequence.

\subsubsection{Effect of aggregation formula}
\begin{table}[h]
\begin{center}
\begin{tabular}{ccc}
	System & Aggregation formula & BLEU \\
	\hline
	\multirow{3}*{T-Base + RTAL(6L)} & Mean  & 27.96 \\
	& Concat + FFN & 28.23 \\
	& EWP + FFN & 28.46	
\end{tabular}
\end{center}
\caption{\label{table4}Impact of aggregation formulas for RTAL. "Mean" denotes aggregation by averaging. "Concat + FFN" denotes aggregation by concatenation. "EWP + FFN" denotes aggregation by element-wise production.}\smallskip
\end{table}

To get better model performance, we try many kinds of aggregation formulas. In this experiment, we evaluate the impact of different choices for aggregation formulas. As shown in table \ref{table4}, the result of averaging layer outputs to fuse information across layers is worse than two other methods, the aggregation by concatenation and aggregation by element-wise production have similar BLEU points. As seen, all the three aggregation formulas efficiently improve the model performance, which shows that our proposed approach is robust.

\subsubsection{Effect of aggregation structure}
To verify the superiority of our proposed approach, we also implement the approaches proposed in the previous works \cite{dou2018exploiting,yu2018deep}. In this experiment, we retrain those different approaches with the same random seed, we also use the same aggregation formula defined in formula \ref{ewp:a} and \ref{ewp:b} for fusing information.

As shown in table \ref{table5}, the linear combination of layers in Transformer outperforms the Transformer only a little, and the iterative combination of layers and CNN-Like-Tree structure in Transformer outperform the baseline model by about 0.3 and 0.4 BLEU points. Although these approaches do not outperform the baseline model significantly, they claim that fusing information across layers is a promising way to improve the model performance. At last, our proposed approach outperform the baseline model significantly, which indicates the superiority of our RTAL in fusing information across layers.

\begin{table}[h]
	\begin{center}
		\begin{tabular}{cccc}
			System & Architecture & BLEU & $\Delta$ \\
			\hline
			\cite{vaswani2017attention} & Transformer-Base & 27.3 & - \\
			\multirow{4}*{Our work} & +Linear Combination & 27.42 & +0.12 \\
			& +Iterative Combination & 27.62 & +0.32 \\
			& +CNN-Like-Tree & 27.76 & +0.46 \\
			& +RTAL (6L) & 28.46 & +1.16
		\end{tabular}
	\end{center}
	\caption{\label{table5} Evaluation of translation performance on WMT14 English-to-German translation task. "+Linear Combination" denotes the linear combination of layers. "+Iterative Combination" denotes the iterative combination of layers. "+CNN-Like-Tree" is a novel structure proposed in the work of \cite{dou2018exploiting}. "+RTAL" denotes residual tree aggregation of layers.}\smallskip
\end{table}

\section{Conclusion}
In this paper, we propose a novel approach to fuse information across layers, which consists of a post-order binary tree and residual connection. To get better model performance, we also try many kinds of aggregation formulas. Our proposed approach is efficient and easy to implement. The experimental results show that our proposed approach outperforms the baseline model significantly.

\bibliographystyle{coling}
\bibliography{coling2020}

\end{document}